# Survey on Deep Learning Techniques for Person Re-Identification Task

Bahram Lavi[1], Mehdi Fatan Serj[2], and Ihsan Ullah[3]

[1]*Institute of Computing, University of Campinas, Sao Paulo 13083-970, Brazil*
[2]*Department of Computer Engineering and Mathematics, University Rovira i Virgili, Tarragona, Spain*
[3]*Data Mining & Machine Learning Group, Discipline of IT, National University of Ireland Galway, Ireland*

## Abstract

*Intelligent video-surveillance* is currently an active research field in computer vision and machine learning techniques. It provides useful tools for surveillance operators and forensic video investigators. Person re-identification (PReID) is one among these tools. It consists of recognizing whether an individual has already been observed over a camera in a network or not. This tool can also be employed in various possible applications such as off-line retrieval of all the video-sequences showing an individual of interest whose image is given a query, and online pedestrian tracking over multiple camera views. To this aim, many techniques have been proposed to increase the performance of PReID. Among the systems, many researchers utilized deep neural networks (DNNs) because of their better performance and fast execution at test time. Our objective is to provide for future researchers the work being done on PReID to date. Therefore, we summarized state-of-the-art DNN models being used for this task. A brief description of each model along with their evaluation on a set of benchmark datasets is given. Finally, a detailed comparison is provided among these models followed by some limitations that can work as guidelines for future research.

## 1 Introduction

The importance of security and safety of people is continuously growing day by day in the society. Governmental and Private organizations are seriously concerned about public areas such as airports and malls. It requires too much expenses and efforts to provide security to the public. To accomplish this goal, video surveillance systems are playing a key role in this manner. Nowadays, plenty of video cameras are growing as a useful tool for addressing various kind of security issues such as forensic investigations, crime preventing, safeguarding the restricted areas, etc.

Continuous recording of videos from network camera per day results in large amount of videos for analysis in a manual video surveillance system. Surveillance operators need to analyze them at the same time for any specific incident or anomaly which is a challenging and tiresome job. Intelligent video surveillance systems (IVSS) aim to automate the issue of monitoring and analyzing the videos from camera networks to help the surveillance operators in handling and understanding the acquired videos by camera networks, which makes it as one of the most active and challenging research area in computer engineering and computer science in which computer vision (CV) and machine learning (ML) techniques are highly required. This field of research enables some various tools such as *on-line* applications for people/object detection and tracking, recognizing a suspicious action/behavior from the camera network; and *off-line* applications to support operators and forensic investigators to retrieve images of the individual of interest from video frames acquired on different camera views.

PReID has been proposed as one of the tools of IVSS. It consists of recognizing an individual over a network of video surveillance cameras with possibly non-overlapping fields of view [1, 2]. Figure 1 shows an example of surveillance cameras with non-overlapping fields of view. Generally speaking, the applications of PReID is to support surveillance operators and forensic investigators in retrieving videos showing an individual of interest, given an image as a query (aka *probe*). Therefore, the video frames or tracks of all the individuals (aka *template gallery*) recorded by the camera network are ranked in order of decreasing similarity to the probe. It allow the user to find the occurrences (if *any*) of the individual of interest hopefully in the top positions.

This is a challenging task due to low image resolution, unconstrained pose, illumination changes, and occlusions which does not allow to exploit strong biometric features like face, etc. Clothing appearance is therefore the most widely used cue; other cues like gait and anthropometric measures have also been investigated. Most of the existing techniques are based on defining a specific descriptor of clothing appearance (typically including color and texture), and a specific similarity measure between a pair of descriptors (evaluated as a *matching score*) which can be either manually defined or learn from data [3, 4, 5, 1, 6]. A standard re-identification method can be described as follows. Let D denotes a descriptor for PReID, $m(\cdot,\cdot)$ the corresponding similarity measure between a pair of images, $\mathbf{T}$ and $\mathbf{P}$ the descriptors of a template and probe image, respectively, and $G = \{\mathbf{T}_1, \ldots, \mathbf{T}_n\}$ the template gallery. Take into consideration that constructing the template gallery depends on a re-identification scenario, which categorized as:
(i) single shot which has only one template frame per individual, and
(ii) multiple shots which contains more than one template frame per individual; in which a continuous re-identification system is employed in real-time where the individual of interest is continuously matched against the template image with the gallery set, using the currently seen frame as probe.
Figure 2 demonstrates a standard person re-identification framework. After image description is generated for probe and each template images of gallery set, the matching scores between each of them is computed; and finally the ranked list is generated by sorting the matching scores in decreasing order.

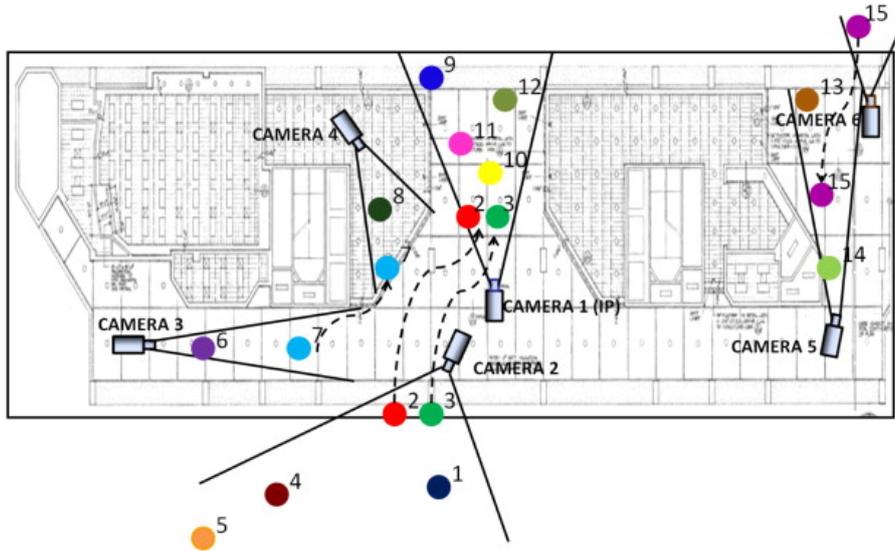

Figure 1: Example of multi-camera surveillance presented for person re-identification problem (figure provided in [7]) .

Apart from the strategy of many descriptors which are mostly based on generating hand-crafted features, recently deep learning networks are specially and widely employed to solve the problem of PReID. Inspired by great success of convolutional neural networks (CNNs) [8] in the area of CV [9, 10], it is therefore adopted by researchers for PReID tasks. Each CNN layer generates a set of feature maps in which each pixel of given image corresponds to a specific feature representation. A desired output is expected at top of the DNN based on the employed model (e.g. classification or Siamese models). The details are presented and discussed in Sect. 3.

This paper presents the state-of-the-art methods about PReID techniques based on deep neural networks (DNN), and provides significant detailed information about them. The literature review involves the papers which are published between year 2014 and 2017. The structure of this paper is organized as follows. In Section 2, the benchmark datasets employed for PReID are briefly explained discussed. Section 3 describes the DNN methods highlighting impact of the key content like objective function, loss functions, data augmentation, etc. Finally, Sect. 5 concludes the paper and gives some future directions for PReID.

## 2 Person Re-identification Benchmark Datasets

To evaluate a PReID method, some factors must be taken into account to reach a reliable recognition rate. This task faces challenges due to occlusions (e.g. apparent on i-LIDS data set) and illumination variation (common in most of them). On the other hand, background and foreground segmentation in order to distinguish person's body is challenging task. Some of the datasets provides already segmented region of interest i.e. the person's body (e.g. VIPeR, ETHZ, and CAVIAR datasets). There are several available datasets that have been prepared to evaluate re-identification task. However, some well-known benchmark data sets like VIPeR, CUHK01, and CUHK03 are mostly used by researchers of this area to evaluate their techniques.

VIPeR is the most challenging due to its challenging images of individual. On the other hand, VIPeR, CAVIAR, and PRID data sets are used when only two fixed camera views are given to evaluate performance of person re-identification methods. Brief description is given on most popular data sets of PReID problem. Additionally, followed by Table 1 with a summary of each dataset.

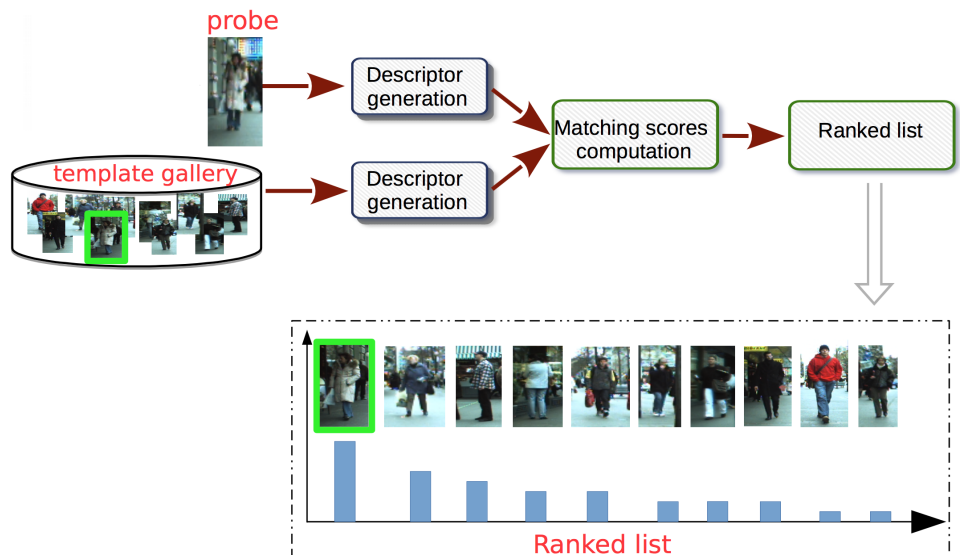

Figure 2: Standard person re-identification system

- **VIPeR** [3]: It is made up of two images of 632 individuals from two camera views with pose and illumination changes. This is one of the most challenging and widely datasets yet for PReID task. The images are cropped and scaled to be $128 \times 48$ pixels.

- **i-LIDS** [11]: It contains 476 images of 119 pedestrians taken at an airport hall from non-overlapping cameras with pose and lightning variations and strong occlusions. A minimum of 2 images and on an average there are 4 images of each pedestrian.

- **ETHZ** [12]: It contains three video sequences of a crowded street from two moving cameras; images exhibit considerable illumination changes, scale variations, and occlusions. The images are of different sizes which can be resized to same width according to the requirements. The data set provides three sequences of multiple images of an individual from each sequence. Sequences 1, 2 and 3 have 83, 35, and 28 pedestrians respectively. Figure 3 shows some example images from VIPeR, i-LIDS, and ETHZ data sets.

- **CAVIAR** [13]: It contains 72 persons and two views in which 50 of persons appear in both views while 22 persons appear only in one view. Each person has 5 images per view, with different appearance variations due to resolution changes, light conditions, occlusions, and different poses.

- **CUHK**: This is provided by Chinese University of Hong Kong. It particularly gathered persons images for person re-identification task, and includes three different partitions with specific set up for each. Figure 4 presents some samples from this data set, including different partitions; *CUHK01* [14] includes $1,942$ images of 971 pedestrians. It has only two images captured in two disjoint camera views, and camera second camera (B) mainly includes images of the frontal view and the back view, and camera A has more variations of viewpoints and poses. Fig. *CUHK02* [15] contains $1,816$ individuals constructed by five pairs of camera views (P1-P5 with ten camera views). Each pair includes 971, 306, 107, 193 and 239 individuals respectively. Each individual has two images in each camera view. This dataset is employed to evaluate the performance when camera views in test are different than those in training. Finally, *CUHK03* [16] includes $13,164$ images of $1,360$ pedestrians. This data set has been captured with six surveillance cameras. Each identity is observed by two disjoint camera views and has an average of 4.8 images in each view; all manually cropped pedestrian images exhibit illumination changes, misalignment, occlusions and body part missing.

- **PRID** [17]: This dataset is specially designed for PReID focusing on single shot scenario. It contains two image sets containing 385 and 749 persons captured by camera A and camera B, respectively. These two datasets share 200 persons in common.

- **WARD** [18]: The dataset has 4,786 images of 70 persons acquired in a real surveillance scenario with three non-overlapping cameras having huge illumination variation, resolution, and pose changes.

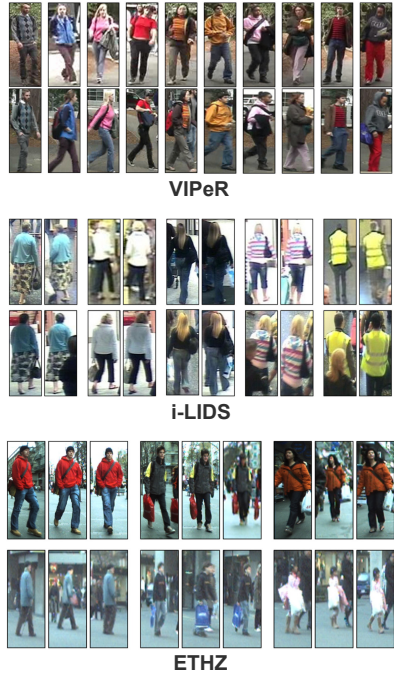

Figure 3: Sample images from VIPeR, i-LIDS, and ETHZ datasets

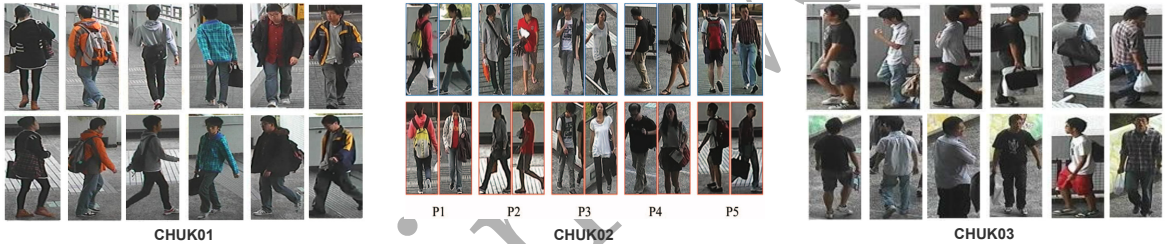

Figure 4: Example of the images from different partition of CUHK data set.

- **Re-identification Across indoor-outdoor Dataset (RAiD)** [19]: It has 6920 bounding boxes of 43 identities captured by 4 cameras. The cameras are categorized in four partitions where the fist two cameras are indoor while the remaining are outdoor. Apparently, the images consists of very large illumination variations because of indoor and outdoor situations.

- **Market-1501** [20]: This is a large PReID dataset which contains 32,643 fully annotated boxes of 1501 pedestrians. Each person is captured by maximum six cameras and each box of person is cropped by a state-of-the-art detector (Deformable Part Model (DPM)) [21].

- **MARS** [22]: It is another large sequence-based PReID dataset which contains 1,261 identities with each identity captured by at least two cameras. It consists of 20,478 tracklets and 1,191,003 bounding boxes.

- **DukeMTMC** [23]: This dataset contains 36,441 manually cropped images of 1,812 persons captured by 8 outdoor cameras. The data set gives the access to some additional information such as full frames, frame level ground truth, and calibration information.

- **MSMT** [24]: This is the most recent and largest PReID dataset. It consists of 126,441 images of 4,101 individuals acquired from 12 indoor and 3 outdoor cameras, with different strong illumination changes, pose, and scale variations.

## 3  Deep Neural Networks for PReID

This section gives an overview of recent works on DNN models for PReID problem. Several interesting DL models have been proposed to improve the performance of PReID either by modifying the existing DL architectures or designing a new DNN. Generally speaking, two types of DL models have been employed in this research area: (i) a classification model for PReID problem, and (ii) a Siamese model based on either pair or triplet comparisons.

Table 1: Summary on benchmark PReID datasets

| Dataset | Year | Multiple images | Multiple camera | Illumination variations | Pose variations | Partial occlusions | Scale variations | Crop image size |
|---|---|---|---|---|---|---|---|---|
| VIPeR | 2007 | | ✓ | ✓ | ✓ | ✓ | | 128 × 48 |
| ETHZ | 2007 | ✓ | | ✓ | | ✓ | ✓ | vary |
| PRID | 2011 | | ✓ | ✓ | ✓ | ✓ | | 128 × 64 |
| CAVIAR | 2011 | ✓ | ✓ | ✓ | ✓ | ✓ | ✓ | vary |
| WARD | 2012 | ✓ | ✓ | ✓ | ✓ | | | 128 × 48 |
| CUHK01 | 2012 | ✓ | ✓ | ✓ | ✓ | ✓ | | 160 × 60 |
| CUHK02 | 2013 | ✓ | ✓ | ✓ | ✓ | ✓ | | 160 × 60 |
| CUHK03 | 2014 | ✓ | ✓ | ✓ | ✓ | ✓ | | vary |
| i-LIDS | 2014 | ✓ | ✓ | ✓ | ✓ | ✓ | ✓ | vary |
| RAiD | 2014 | ✓ | ✓ | ✓ | ✓ | | | 128 × 64 |
| Market-1501 | 2015 | ✓ | ✓ | ✓ | ✓ | ✓ | ✓ | 128 × 64 |
| MARS | 2016 | ✓ | ✓ | ✓ | ✓ | ✓ | | 256 × 128 |
| DukeMTMC | 2017 | ✓ | ✓ | ✓ | ✓ | | ✓ | vary |
| MSMT | 2018 | ✓ | ✓ | ✓ | ✓ | | ✓ | vary |

DL models in PReID problem still suffer from the lack of training data samples. This is due to the reason that majority of the PReID datasets provide only two images per each individual (e.g. VIPeR dataset [3]) that makes the model fail at test time due to overfitting. Therefore, a pairwise Siamese network has been proposed for this task [16]. The models take pair of images as individual input to each neural network in a Siamese model. Employing a Siamese model could be a good solution to train on existing PReID datasets[25] having fewer training samples. In coming subsections, state-of-the-art DNNs will be critically discussed followed by brief discussion on essential contents of DNN such as loss functions and data augmentation technique.

## 3.1 Models based on Classification

In person re-identification task, typically, a model based on classification problem requires determining the individual identity (aka *class*) that it belongs to. Figure 5 presents a standard CNN-based model; in which the network takes the image of an individual, and computes probability of corresponding class of the individual. For the data sets with less instances per individual (e.g. VIPeR), the classification models can be easily failed due to overfitting problem.

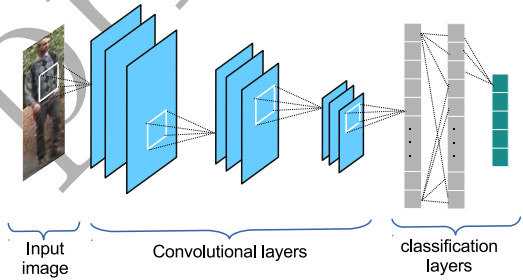

Figure 5: An example of a standard classification model.

Wu et al. [26] proposed a feature fusion deep neural network to regularize CNN features jointly by hand-crafted features. The network takes a single image size of $224 \times 224 \times 3$ as the input of the network, and on the other hand, the hand-crafted features are extracted by one of the standard person re-identification descriptor from that image (best performance obtained from ELF descriptor [27]). Then, both extracted features are followed by a buffer layer and a fully connected layer which are acting as the fusion layer. The buffer layer is used for the fusion action which is essential since it bridges the gap between two features with huge difference(i.e. hand-crafted features and deep features). A softmax loss layer then takes the output vector of fully connected layer in order to minimizing the cross-entropy loss, and outputs the deep feature representation(the output size of the softmax layer is varied for different kind of hand-crafted features). The whole network is trained by applying mini-batch stochastic gradient for back propagation. Note that the parameters of the whole CNN network are effected by the hand-crafted features. They implemented the model by employing the deep learning caffe framework and replaced the last three layers of this framework with buffer layer, fully connected layer, and softmax layer. In [28], low-level descriptors including SIFT and color histograms are extracted from the LAB color space over a set of 14 overlapping patches in size of $32 \times 32$ pixels with 16 pixels of stride. Then the dimensionality of SIFT and color histogram features are reduced with PCA. Then, those features are further embedded using Fisher vector encoding in order to producing feature representations which are linear separable. Note that one Fisher vector is computed on the SIFT and another one on the color histogram features, and finally two fisher vectors are concatenated in a single vector. A hybrid network builds fully connected layers on the input of Fisher vectors and employs the linear discriminative analysis (LDA) as an objective function in order to maximizing margin between classes.

Xiao et al. [29] proposed learning deep features representations from multiple data sets by using CNNs to discover effective neurons for each training data set. They first produced a strong baseline model that works on multiple data sets simultaneously by combining the data and labels from several re-id data sets together and trained the CNN with a softmax loss. Next, for each data set, they performed the forward pass on all its samples and compute for each neuron its average impact on the objective function. Then, they replaced the standard Dropout with the deterministic Domain Guided Dropout in order to discarding useless neurons for each data set, and continue to train the CNN model for several more epochs. Some neurons are effective only for a specific data set which might be useless for another one, this caused by data set biases. For instance, the i-LIDS is the only data set that contains pedestrians with luggage, thus the neurons that capture luggage features will be useless to recognize people from the other data sets.

In [30], a deep convolutional model proposed in order to handle misalignments and pose variations of pedestrian's images. The overall multi-class person re-identification network composed by two sub-network; first a con convolutional model is adopted to learn global features from the original images. Then a part-based network is used to learn local features from an image which includes six different parts of pedestrian body. Finally, both sub-networks are combined following by a fusion layer as the output of the network, with the shared weight parameters during training process. The output of the network is further used as an image signature to evaluate the performance of their person re-identification approach with Euclidean distance. The proposed deep architecture explicitly enables to learn effective feature representations on the person's body part and adaptive similarity measurements.

Li et al. [31] designed a multi-scale context aware network to learn powerful features throughout the body and body parts, which can capture knowledge of the local context by stacking convolutions of multiple scales in each layer. In addition, instead of using predefined rigid parts, is proposed to learn and locate deformable pedestrian parts through networks of spatial transformers with new spatial restrictions. Because of variations and background disorder raised in some difficulties in representation based on parts, the learning processes of full-body representation integrated with body parts into a unified framework for multi-class identification.

### 3.2 Models based on Siamese Network

As pointed out, Siamese network models have been widely employed in person re-identification task due to lack of training instances on this area research. Siamese neural network is a type of neural network architectures which contains two or more identical sub-networks; it is worth to point out that here identical means the sub-networks share same network architecture and the same parameters and weights for each other(aka *shared weight parameters*, which is usually indicated by w between each sub-networks). A Siamese network can be typically employed as pairwise: when two sub-networks included, or triplet: when three sub-networks. The output of Siamese model is basically a similarity score which takes place at the top of the network. Let $X = \{x_1, x_2, \ldots, x_n\}$ and $Y = \{y_1, y_2, \ldots, y_n\}$ be a set of person images and corresponding label for each, respectively. and to distinguish from the positive and negative pairs

$$I_s(x_i, x_j) = \begin{cases} positive & if \quad y_i = y_j, \\ negative & if \quad y_i! = y_j \end{cases} \quad (1)$$

For triplet Siamese model an objective function is used to train the network models, which creates a margin between distance metric of positive pair and distance metric of negative pair. In order to output for this type of Siamese model, a softmax layer is employed at the top of the network on both distance outputs. The triplet loss function is used to train the network models, which makes the distance between the matched pairs less than the mismatched pairs in the learning feature space. Let $O_i = \{(I_i, I_i^+, I_i^-)\}_{i=1}^N$ be a set of triplet images, in which $I_i$ and $I_i^+$ are referred to images of the same person, and $I_i$ and $I_i^-$ present the different persons. Figures 6 and 7 present an example of Siamese models for pairwise and triplet comparison, respectively.

#### 3.2.1 Pairwise models

In [32], a Siamese pair-based model takes two images as the input of the two sub-networks which are locally connected to the first convolutional layer. They employed a linear SVM at top of the network instead of using the softmax activation function in order to measure similarity of input images pair as the output of the network. In [33], a Siamese neural network has been constructed to learn pairwise similarity. Each input image of pair first partitioned into three overlapping horizontal parts. The part pairs are matched through three independent Siamese networks, and finally are fused at the score level. Li et al. [16] proposed a deep filter pairing neural network to encode photo-metric transformation across camera views. A patch matching layer is further added to the same network in order to multiply the convolution feature maps of pair images in different horizontal stripes. Later, Ahmed et al. [34] improved the pair-based Siamese model in which the network takes pair of images as the input, and outputs the probability of whether two images in the pair is referred to the same person or different people. The model begins with two layers of convolution by passing input pair of images. The generated feature maps are passed through a max-pooling kernel to another convolution layer followed by another max-pooling layer in order to decreasing the size of feature map. Then a cross-input neighborhood layer computes the differences of the features in neighboring locations of the other image.

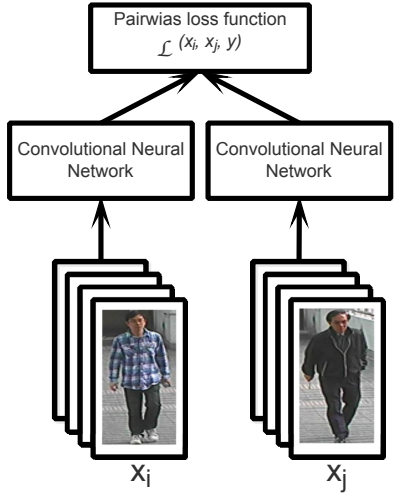

Figure 6: An example of a pairwise Siamese model which takes pair of images as its input.

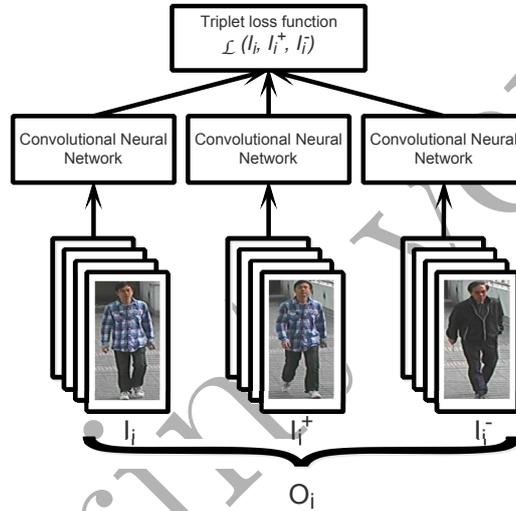

Figure 7: An example of a triplet Siamese model which takes three images as its input.

Wang et al. [35] designed an ensemble of multi-scale and multi-part with CNNs to learn image representations and similarity measure jointly. The network takes two person images as input of the network and derived the full scale, half scale, top part and middle part image pairs from the original images. The network outputs the similarity score of the pair images. This architecture is composed of four separate sub-CNNs with each sub-CNN embedding images of different scales or different parts. The first sub-CNN takes full images of size $200 \times 100$ and the second sub-CNN takes down-sampled images of size $100 \times 50$. The next two sub-CNNs take the top part and middle part as input respectively. Four sub-CNNs are all composed of two convolutional layers, two max pooling layers, a fully connected layer and one L2-normalization layer. They obtained the image representation from each sub-CNN and then calculate their similarity score. The final score is calculated by averaging four separate scores; ReLU activation function is used as the neuron activation function for each layer, Dropout layer is used in the fully connected layer to reduce the risk of overfitting problem. Wang et al. [36] developed a CNN model in order to jointly learn single-image representation (SIR) and cross-image representation (CIR) for person re-identification task. Their approach relied on investigating on two separate models for comparing pairwise and triplet images (is explained at the next section) with similar deep structure. Each of these models consists of different sub-networks for SIR and CIR learning, and a sub-network shared by SIR and CIR learning. For the pairwise comparison, they used a Euclidean distance as loss function to learn SIR, and formulated the CIR learning as a binary classification and employed the standard SVM to learn CIR as its loss function. They used the combination of both loss functions as the overall loss function of pairwise comparison. Moreover, for the triplet comparison, the loss function to learn SIR makes the distance between the matched pairs less than the mismatched pairs. The CIR learning formulated as a learning-to-rank problem and employed the RankSVM as its loss function, and finally, the combination of both are used as the overall loss function of triple comparison. The shared sub- network used with sharing parameters during the training stage.

Liu et al. [37] utilized a deep learning model to integrate a soft attention based model in a Siamese network. The model focuses on the important local parts of input images on pair-based Siamese model. Chen et al. [38] proposed a deep ranking framework to jointly learn joint representation and similarities for comparing pairwise images. To this aim, a deep CNN is trained in order to assign a higher similarity score to the positive pair than any negative pairs in each ranking unit by utilizing the logistic activation function, which is employed as $\sigma(x) = log_2(1 + 2^{-x})$. They first stitch pair of persons images horizontally to form of an image which used as the input of the network, and then, the network returns a similarity score as its output. Franco et al. [39] proposed a coarse-to-fine approach to achieve a generic-to-specific knowledge through a transfer learning. The approach is followed by three steps: first a hybrid network is trained to recognize a person, then another hybrid network employed to discriminate the gender of person; finally the output of two networks are passed through the coarse-to-fine transfer learning method to a pairwise Siamese network to accomplish the final person re-identification goal in terms of measuring the similarity between those two features. Later, the same authors proposed a novel type of features based on convolutional covariance descriptor (CCF) in [40]. They intend to obtain a set of local covariance matrices over the feature maps extracted by the hybrid network under the same strategy of their former method.

Wang et al. [41] proposed a pairwise Siamese model by embedding a metric learning method at the top of the network to learn spatio-temporal features. The network takes a pair of images in order to obtain CNN features, and outputs whether two images belong to the same person or different one by employing the quadratic discriminant analysis method. In [42], a Siamese network takes a CNN learning feature pair, and outputs the similarity value between them by applying the cosine/Euclidean distance function. A CNN framework employed to obtain deep features of each input image pair, and then, each image is split into three overlapping color patches. The deep network built in three different branches and each branch takes a single patch as its input; finally, the three branches are concluded by a fully-connected layer.

The multi-task deep learning proposed in [43] aimed to use a separate soft-max for each auxiliary task as identification, pose labeling, and each attribute labeling task. A CNN is used to generate image representation in which a single image of size $64 \times 64$ used as the input. The model proposed at this work is pairwise model; which is designed a multi-task learning which takes a pair of images by embedding a specific cost function for each. For instance, they used a softmax regression cost function for each of individual task in which a multi-class linear classifier calculates the probability of its input belongs to each class. They minimized the cost function using stochastic gradient descent with respect to weights of each task; then, the linear combination over all the cost functions is presented as the final cost function of the network. The designed network is composed of three convolutional layers, and two max-pooling layers, followed by a final fully connected layer as the output of network. The hyperbolic tangent activation function was used between each convolutional layer, while a linear layer was used between the final convolutional layer and the fully connected layer. The activation of neurons in the fully connected layer gives the feature representation of the input image, dropout regularization was used between the final convolutional layer and the fully connected layer. A Strict Pyramidal Deep Metric Learning approach is proposed in [44], in which the Siamese network composed by two strict pyramidal CNN blocks with shared parameters between each, and produced salient features of an individual as the output of the network. The authors was willing to present a simple connection scheme to trade-off between lower computational and memory costs with respect to the others NNs.

Similar to Siamese networks, a Deep Hybrid Similarity Learning (DHSL) [45] based on CNN model is proposed to learn similarity between pair of images. In their approach, a two-channel CNN with ten layers aims to learn pair feature vectors to discriminate input pair in order to minimizing network output value for similar pair images and maximizing from different ones. A new hybrid distance method using element-wise absolute difference and multiplication proposed to improve CNN in similarity metrics learning. A novel scheme, called deep multi-view feature learning (DMVFL), proposed in [46] aimed to combine handcrafted features (e.g. LOMO [47]) with generated deep features using by CNN model; and embedding a metric distance method at top of the network in order to learn metric distance. To this aim, the XQDA metric learning method [47] utilized regarding to jointly learning of handcrafted and deep learning features; in this manner, it is possible to investigate how handcrafted features could be influenced by deep CNN features. A two-channel convolutional neural network by new component named Pyramid Person Matching Network (PPMN) proposed in [48] with the same architecture of GoogLeNet; in which the network takes a pair of images as input, and extracts semantic features by convolutional layers; and finally the Pyramid Matching Module aimed to learn corresponding similarity between semantic features based on multi-scale convolutional layers. Two fully-connected layers and softmax last layer at the top of the network are used to distinguish whether pair images belongs to same person or not.

A novel Siamese Long-Short Term Memory (LSTM) based architecture was proposed in [49] which aims to lever contextual dependencies by selecting the relevant contexts in order to enhance the discriminative capabilities of the local features. They proposed a pairwise Siamese model which contains six LSTM models within each subnetwork. First, each individual image is divided into six horizontal non-overlapping parts, and from each part an image representation is extracted by using two state-of-the-art descriptors (i.e. LOMO and Color Names). Then, each feature vector is separately fed to an individual LSTM network with the share parameters. The outputs from each LSTM network are combined, and the relative distance of subnetworks is computed by Contrastive loss function. The whole pairwise network is trained with mini-batch stochastic gradient descent algorithm. Qian et al. [50] proposed a multi-scale model which is able to learn discriminant feature from multi-scales in terms of different levels of resolution. A saliency-based learning strategy is adopted to learn weight important scales. In parallel with the pairwise Siamese model which aims to distinguish whether a pair of images is belong to the same person or not, a tied layer is also used between each layers of each branch in order to verifying the identity of an individual. The designed model consists of five components: tied convolutional layers, multi-scale stream layers, saliency-based learning fusion layer, verification subnet and classification subnet.

### 3.2.2 Triplet models

Ding et al. [51], proposed a deep CNN framework in order to produce feature representation from a raw person images. Each single triplet images unit is used as the input of the network of size $250 \times 100$ pixels aiming to maximize the relative distance between pair of images of the same person and different person under $L_2$ loss function. Each triplet training sample is separately fed into three identical networks with the shared parameter set between them, and train the network based on gradient descent algorithm with respect to the output feature of the network.

Zhang et al. [52] presented a novel supervised formulation for the tasks of general image retrieval and person re-identification across disjoint camera views, jointly by feature learning and hash function learning via deep neural networks. The deep architecture of neural network utilized to produce the hashing codes with the weight matrix by taking raw image size of $250 \times 100$ for person re-identification datasets as input of the network. The objective of the hash code is looked for a single projection to be mapped each sample into a binary vector. The network is used to be trained under triplet-based similarity learning to enforce that the images of similar persons should have similar hash codes. For each triplet unit, they organized to maximize the margin between the matched pairs and the mismatched pairs. The deep architecture of neural network utilized at this work is the Alexnet pre-trained network which consists of ten layers: the first six layers form the convolution-pooling network with rectified linear activation and average pooling operation. They used 32, 64, and 128 kernels with size $5 \times 5$ in the first, second, and third convolutional layers and the stride of 2 pixels in the every convolution layer. The stride for pooling is 1 and they set the pooling operator size as $2 \times 2$. The last four layers is compromised with two standard fully-connected layers, and tangent like layer to generate the output as the hash codes, and an element-wise connected layer to manipulate the hash code length by weighting each bin of the hashing codes. The number of units set 512 in the first fully-connected layer and the output of the second fully-connected layer equals to the length of hash code. The activation function of the second fully-connected layer is the tanh-like function, while ReLu activation function is adopted for the others. Cheng et al. [53] proposed a triplet loss function in which the network takes a triplet unit of images as its input, and enables the network to be jointly learnt from the global full-body and local body-parts features; the fusion of these two types of features at the top of the network is presented as the output of the network. The utilized CNN model begins with a convolution layer which each of them is divided into four equal parts, and each part forms the first layer of an independent body-part channel aiming to learn features from the representative body part. The four body-part channels together with the full-body channel constitute five independent channels that are trained separately from each other (with no parameter sharing between each). At the top of the network, the outputs obtained from five separate channels are concatenated into a single vector, and is passed through a final fully-connected layer. Su et al. [54] proposed a semi-supervised three-stage learning approach in which the network foremost is trained on an independent data set to predict some predefined attributes, and the distinguished attributes are further trained on other data sets with individual class labels.

Liu et al. [55] proposed multi-scale triplet network copmpromised with one deep convolutional neural network and two shallow neural networks (i.e in order to produce less invariance and low-level appearance features from images), with shared parameters between them. The deep network designed with five convolutional layers, five max-pooling layers, two local normalization layers, and three fully-connection layers, while each shallow network composed by two convolutional layers followed by two pooling layers. The output of each network are further fused at an embedding layer in order to generate final feature representation.

A structured graph Laplacian algorithm is utilized in [56] within a CNN model. Despite from traditional contrastive and triplet loss in terms of joint learning, the structured graph Laplacian algorithm is additionally embedded at the top of the network; the designed network jointly learns within a triplet Siamese network, where the softmax function is used to maximize the inter-class variations of different individual, while the structured graph Laplacian algorithm is employed to minimize the intra-class variations. As the authors pointed out, the designed network needs no additional network branch which makes the training process more efficient.

Bai et al. [57] proposed a deep-person model in order to generate global-based and body-part-based feature representation of persons image. Each image of triplet unit is fed into a backbone convolutional neural network to generate low-level features with the shared parameters. The output features of the backbone network are further fed into a two-layer Bidirectional LSTM aiming to generate part-based feature respresentations; LSTM is adopted because of its discriminative ability of part representation with contextual information, handlign the misalignment with the sequence-level person representation. At the same time, the output features are also fed into another network branch which compromised with a global average pooling, a fully-connected, and a Softmax layer for global feature learning. Finally, the output features are leanrt distances under a triplet loss function by adopting another branch of network during the trainign of the whole network. A coherent and conscious deep learning approach was introduced in [58] which ables to cover whole network cameras. The proposed approach aims to seek the globally optimal matching over the cameras. The deep features are generated over full body and part body under a triplet framework, in which each image within a triple unit is presented a sample image of one camera view, while the other images are presented from other camera views. Once the deep features generated, the cosine similarity is used to obtain the similarity scores between them, and afterward, the gradiant descent is adopted to obtain the final globally optimal association. All the calculations are involved in both forward and backward propagation to update CNN features.

## 3.3 Loss Functions for PReID

In most of the statistical areas such as ML, computational neuron-science, etc., a loss function (*aka cost function*), aims to map intuitively some values into a one single real number. This represents a cost which is associated to those values. The techniques like NNs are in the same way to optimally minimize that loss function. When, a loss function is used for a Siamese model, it depends on the type of model which is going to be chosen (i.e. pairwise or triplet model). In follwing subsections, we discuss loss functions that are commonly used on pair- and triplet-based models for PReID.

### 3.3.1 Pairwise loss function:

The most common loss functions used are given below.

- **Hinge loss**: This loss function basically refers to maximum-margin classification; the output of this loss is become zero when the distance similarity of the positive pairs is greater than the distance of the negative ones with respect to the margin $m$. This loss is defined as follow

$$I(x_1, x_2, y) = \begin{cases} \|x_1 - x_2\| & if \quad y = 1 \\ max(0, m - \|x_1 - x_2\|) & if \quad y = -1 \end{cases} \quad (2)$$

- **Cosine similarity loss**: This similarity loss function maximizes the cosine value for positive pairs and reduce the angle in between them, and at the same time, minimize the cosine value for the negative pairs when the value is less than margin.

$$I(x_1, x_2, y) = \begin{cases} max(0, cos(x_1, x_2) - m) & if \quad y = 1 \\ 1 - cos(x_1, x_2) & if \quad y = -1 \end{cases} \quad (3)$$

- **Contrastive loss** [59]: This loss function minimizes meaningful mapping from high to low dimensional space maps by keeping the similarity of input vectors of nearby points on its output manifold and dissimilar vectors to distant points. Accordingly, the loss can be computed as:

$$I(x_1, x_2, y) = (1 - y)\frac{1}{2}(D)^2 + (y)\frac{1}{2}\{max(0, m - D)\}^2 \quad (4)$$

  where $m > 0$ is a margin parameter acting as a boundary, and $D$ is a distance between two feature vector that is computed as $D(x_1, x_2) = \|x_1 - x_2\|_2$.

The average of total loss for each above-mentioned pairwise loss functions is computed as:

$$\mathcal{L}(X_1, X_2, Y) = -\frac{1}{n}\sum_{i=1}^{n} I(x_i^1, x_i^2, y_i) \quad (5)$$

### 3.3.2 Triplet loss function

This loss function basically creates a margin between distance metric of positive pair and distance metric of negative pair. The triplet loss function is used to train the network models, which makes the distance between the matched pairs less than the mismatched pairs in the learning feature space. Let $O_i = \{(I_i, I_i^+, I_i^-)\}_{i=1}^N$ be a set of triplet images, in which $I_i$ and $I_i^+$ are referred to images of the same person, and $I_i$ and $I_i^-$ present the different persons.

**Euclidean distance** is commonly used as the distance metric of this function. The loss function under $L2$ distance metric has been employed in some of the triplet-based models such as [51, 37, 53, 54], and is denoted as $d(W, O_i)$; where $W = W_i$ is the network parameters, and $F_w(I)$ represents the network output of image $I$. The difference in the distance is computed between the matched pair and the mismatched pair of a single triplet unit $O_i$:

$$d(W, I_i) = \|F_W(I_i) - F_W(I_i^+)\|^2 - \|F_W(I_i) - F_W(I_i^-)\|^2 \tag{6}$$

**Hinge loss** function aims to minimize the squared hinge loss of the linear SVM which is equivalent in order to finding the max margin according to the true person match and false person match over training step. This loss function is a convex approximation in range of 0-1 ranking error loss, which approximate the model's violation of the ranking order specified in the triplet.

$$\mathcal{L}(I_i, I_i^+, I_i^-) = max(0, g + D(I_i, I_i^+) - D(I_i, I_i^-)) \tag{7}$$

where $g$ is a margin parameter that regularizes the margin between the distance of the two image pairs: $(I_i, I_i^+)$ and $(I_i, I_i^-)$, and $D$ is the euclidean distance between the two euclidean points. An improved triplet loss function is:

$$\mathcal{L}(I_i, I_i^+, I_i^-, w) = \frac{1}{N} \sum (max\{d^n(I_i, I_i^+, I_i^-, w), \delta_1\} + \beta max\{d^p(I_i, I_i^+, I_i^-), \delta\_2\}), \tag{8}$$

where $N$ is the number of triplet training examples, $\beta$ is a weight to balance the inter-class and intra-class constraints; in which the distance function $d(.,.)$ is defined as the L2-norm distance:

## 3.4 Activation functions

For each neuron within a neural network a relative output is provided and determined by a specific predefined activation function on its inputs. Among different types of the activation functions, non-linear activation functions is quite common to approximate non-linear functions. This is typically calculated among the all neurons, starting from the neurons of beginning layers to neurons of output layers. In the task of person re-identification, most of the works utilized *ReLu* activation function. Only in [52, 60, 43, 61, 44], the authors attempted to use *Hyperbolic-tangant* function.

## 3.5 Optimizers

In the task of person re-identification, most of the discussed works employed Stochastic Gradient Descent (SGD) to train the network. Only in [60, 49] and [36], they adopted RMSProp [62] and RankSVM [63] for training the network, respectively. The authors in [60] claimed that RMSProp, as an adaptive gradient decent algorithm, is more suitable to deep layers, and converges much faster than SGD. For more details, we refer the reader to the reference paper as indicated above.

## 3.6 Data Augmentation

Data augmentation is crucial mechanism for training a neural network which must be carefully taken into account before train a network. Regarding to the lack of training data, which leads to lack of positive samples (matched pairs) and negative samples (non-matched pairs), the network usually fails in overfitting problem; to alleviate the risk of overfitting, the data augmentation is proceed in order to generating artificial samples to increase the number of training samples, which increase the diversity of network during training at the same time. In person re-identification, this typically is done by a simple transformation of an individual image of size $H \times W$ into a several number of uniform distribution in range of $[-\alpha H, \alpha H] \times [-\alpha W, \alpha W]$ for small random of perturbation per each individual image.

Table 2: Comparison of existing DL models based on Rank-1 recognition rates PReID. Type of models (pairwise, triplet, and classification) are denoted by $P_s$, $T_s$, and $C$ and colored by red, green and blue, respectively. This table is best viewed in color.

| Ref.# | Year | Model | VIPeR | CUHK01 | CUHK03 | i-LIDS | PRID-2011 | CAVIAR | MARS | Market-1501 | WARD |
|---|---|---|---|---|---|---|---|---|---|---|---|
| Li [16] | 2014 | $P_s$ | — | 20.65 | — | — | — | — | — | — | — |
| Zhang [32] | 2014 | $P_s$ | 12.50 | — | — | — | — | 7.20 | — | — | — |
| Yi [33] | 2014 | $P_s$ | — | 28.23 | — | — | — | — | — | — | — |
| Ahmed [34] | 2015 | $P_s$ | 34.81 | 65.00 | 54.74 | — | — | — | — | — | — |
| Ding [51] | 2015 | $T_s$ | 40.50 | — | — | 52.10 | — | — | — | — | — |
| Zhang [52] | 2015 | $T_s$ | — | — | 18.74 | — | — | — | — | — | — |
| Shi [64] | 2015 | $P_s$ | 40.91 | 86.59 | 59.05 | — | — | — | — | — | — |
| Liu [37] | 2016 | $P_s$ | — | 81.40 | 65.65 | — | — | — | — | — | — |
| Cheng [53] | 2016 | $T_s$ | 47.80 | 53.70 | — | — | 22.00 | — | — | — | — |
| Chen [38] | 2016 | $P_s$ | **52.85** | 57.28 | — | — | 66.62 | **53.60** | — | — | — |
| Wu [26] | 2016 | $C$ | 51.06 | 55.51 | — | — | — | — | — | — | — |
| Xiao [29] | 2016 | $C$ | 38.60 | 66.60 | 75.30 | 64.60 | 64.00 | — | — | — | — |
| Wu [60] | 2016 | $P_s$ | — | 71.14 | 64.90 | — | — | — | — | 37.21 | — |
| Li [65] | 2016 | $P_s$ | — | — | — | — | — | — | — | 59.56 | — |
| Shi [42] | 2016 | $P_s$ | 40.91 | 69.00 | — | — | — | — | — | — | — |
| Varior [49] | 2016 | $P_s$ | 42.40 | — | 57.30 | — | — | — | — | — | — |
| Wang [35] | 2016 | $P_s$ | 40.51 | 57.02 | 55.89 | — | — | — | — | — | — |
| Wang [36] | 2016 | $P_s$ | 29.75 | 58.93 | 43.36 | — | — | — | — | — | — |
| Wang [36] | 2016 | $T_s$ | 35.13 | 65.21 | 51.33 | — | — | — | — | — | — |
| Franco [39] | 2016 | $P_s$ | — | 44.94 | 63.51 | 62.30 | 53.33 | — | — | — | — |
| Wu [28] | 2016 | $C$ | 44.11 | 67.12 | — | — | — | — | — | 48.15 | — |
| Wang [41] | 2016 | $P_s$ | — | 38.28 | 27.92 | — | — | — | — | — | — |
| Su [54] | 2016 | $T_s$ | 43.50 | — | — | — | 22.60 | — | — | — | — |
| Mclaughlin [43] | 2016 | $P_s$ | 33.60 | — | — | — | — | — | — | — | — |
| McLaughlin [61] | 2016 | $P_s$ | — | — | — | **85.00** | **70.00** | — | — | — | — |
| Liu [55] | 2016 | $T_s$ | — | — | — | — | — | — | — | 55.40 | — |
| Iodice [44] | 2016 | $P_s$ | 18.04 | — | — | — | — | — | — | — | — |
| Su [30] | 2017 | $C$ | 51.27 | — | 78.29 | — | — | — | — | 63.14 | — |
| Franco [40] | 2017 | $P_s$ | — | 63.85 | 63.90 | 55.85 | — | — | — | — | — |
| Qian [50] | 2017 | $P_s$&$C$ | 43.30 | 79.01 | 76.87 | 41.00 | 65.00 | — | — | — | — |
| Zhu [45] | 2017 | $P_s$ | 44.87 | — | — | — | — | — | — | — | — |
| Cheng [56] | 2017 | $T_s$ | — | 70.09 | 84.70 | — | — | — | — | 83.6 | — |
| Tao [46] | 2017 | $P_s$ | 46.00 | — | — | — | — | — | — | — | — |
| Mao [48] | 2017 | $P_s$ | 45.82 | **93.10** | **85.50** | — | — | — | — | — | — |
| Li [31] | 2017 | $C$ | 38.08 | — | 74.21 | — | — | — | **71.77** | 80.31 | — |
| Lin [58] | 2017 | $T_s$ | — | — | — | — | — | — | — | 81.15 | 99.71 |
| Bai [57] | 2017 | $T_s$ | — | — | **91.50** | — | — | — | — | **92.31** | — |
| Chung [66] | 2017 | $T_s$ | — | — | — | 60.00 | **78.00** | — | — | — | — |

# 4 Comparison and Open Issues

**Performance Measure** To evaluate the performance of a PReID system, the cumulative matching characteristic (CMC) curve is typically calculated and demonstrated as a standard recognition rate of which the individuals are correctly identified within a sorted ranked list. On the other words, a CMC curve is defined as the probability that the correct identity is within the first $'r'$ ranks, where r = 1, 2, ..., n, and $n$ is the total number of template images involved during the testing of a PReID system. By definition, the CMC curve increases with $'r'$, and eventually equals 1 for r = n. The authors of this work attempted to collect the original CMC curves presented at each of discussed works within this survey for sake of a comprehensive comparisons. However, the CMC curves of most of those works neither are publicly available, nor even responded to our contacts, we therefore compare only the first-rank recognition rate of existing deep PReID techniques. To compare the performance of the methods with each other, we report the first rank (Rank-1) accuracy in table 2. Rank-1 has higher importance in PReID due to the reason that the system needs to recognize the person from the limited hard to recognize available data in first glance.

We combined, summarized, and compared different DL methods for PReID since 2014 until 2017. These are shown in Table 2. They are compared based on their Rank-1 recognition rate performance over specific PReID datasets. The type of models that are pairwise, triplet, and classification are denoted by $P_s$,$T_s$,and $C$ and colored by red, green and blue, respectively. The global best results among the methods is shown in bold. Particularly, the best result with respect to the model type is indicated by its corresponding color (e.g. the best result among the pairwise models is colored by red). Further, we will discuss and highlight the best methodology and combination of training algorithm with loss function and optimizer to attain significant performance in PReID.

In case of VIPeR and CAVIAR datasets, the deep CNN model proposed by Chen et al. [38] outperformed others in performance. The distance between image pairs is computed with Logistic function. Authors believe that stitching the pair of pedestrian images at the beginning enables the architecture to be robust even on datasets with lack of image samples per individual. Moreover, the good performance of proposed method comes not only from the deep representation learning but also the ranking algorithm because of the basic differences between image classification and ranking tasks. The learning of ranking features algorithm is based on a relative similarity rather than an absolute similarity value for each pair, which showed better result for PReID.

On CUHK01, the method proposed by Mao et al. [48] outperformed other models using their proposed pyramid matching strategy. This shows its robustness and effectiveness on addressing the misalignment and variation issues posed by different viewpoint. A Softmax layer is used to distinguish the similarity between image inputs.

CUHK03 is used by the model proposed by Bai et al. [57] which shows 6% higher Rank-1 i.e. 91.50% than previous best performance by [48]. The same model also showed more than 10% higher rate as compare to Cheng et al. [56] on Market-1501 dataset. This Deep-Person approach, in order to learn highly discriminative deep features, jointly focus on feature representation and feature learning by taking into consideration of complementary advantages on both aspects. It uses a triplet Siamese network that estimate its cost using Euclidean distance as the loss function.

The pairwise Siamese network in [61] outperformed other models on i-LIDS dataset by achieving 85%. This model uses a recurrent neural network architecture proposed for video sequences of individuals. Each sub-network takes three sequences frame of a person and passed them though CNN and RNN. The Siamese cost of each sub-network is computed by the Euclidean distance (i.e. the similarity between network pair). The Hyperbolic-tangent was used as the activation function. The proposed architecture in [66] achieved the best Rank-1 recognition of 78% for PRID-2011 dataset as compared to 70% by [61]. This model is able to learn spatio-temporal information separately and as a whole to handle viewpoint and pose invariants. The model computes the similarity between images of triplet units by Euclidean distance and using a hyperbolic-tangent function as an activation function. Finally, only the works in [31] and [58] have evaluated their methodology on MARS and WARD datasets, respectively. The Rank-1 rate on MARS i.e. 71.7% shows and leave further space for future research, however, the recognition rate in [58] for WARD dataset shows almost optimal performance leaving little margin for future research. However, still in surveillance, once need 100% recognition rate to avoid the anomalies.

All the model discussed here have trained their model with *SGD* with back-propagation algorithm. Majority of these works evaluated their deep models for PReID on CUHK01, CUHK03, VIPeR, i-LIDS, PRID, and Market-1501 datasets. Table. 2 shows that VIPeR dataset is used mostly in PReID problem but it still remains one of the most challenging dataset.

Good performances have been shown in various large models, however, in real scenarios the models need to be fast and effective. Almost in many video surveillance system, the concept of processing time is neglected for the sake to achieve higher accuracy. However, it should always be taken into consideration since it is very costly due to the requirement of powerful computers to run these deep models. Efforts have to be made in this regard to make methods more efficient and compatible for achieving high performance despite smaller size of the network [44]. The network can be reduced by either reducing the number of layers, number of parameters, or introducing a new scheme that has lower connectivity. In [67, 68], authors aimed to trade-off between ranking accuracy and processing time by proposing a multi-stage ranking system and showed promising results.

Additionally, the task of PReID still suffers from the lack of training data samples. Although, this problem is addressed with the help of pairwise Siamese networks which have shown promising performance. However, large scale datasets are needed to make models more reliable to tackle challenges such as pose and viewpoint variations in the images.

In machine learning, a classification problem can be more often adopted to the problems with limited number of classes in which massive number of instances per class are highly demanded. To this end, the existing methods of machine learning such as artificial neural networks allow to solve classification problems with above-mentioned limitations. In PReID, number of persons and the corresponding classes are increasing day by day. However, the number of instances acquired from the camera networks are very limited. In this manner, PReID can not be perfectly taken in a position as a standard classification problem, particularly with DNNs. In contrary with traditional classification problem, metric learning methods, as discussed in this paper, can help and overcome the limitation of deep models as an appropriate tool for solving PReID problem.

## 5 Conclusion

PReID is one the most challenging task for intelligent video surveillance system with open application areas in numerous fields. Despite high importance, still it is facing problems due to poor performance of the models in the real world scenarios. In this survey, we summarized the recent advances being done with DNNs for PReID task since 2014 until 2017. We have shown the type of models and widely used activation and loss functions for PReID. In addition, we highlighted all the available datasets in this domain. VIPeR dataset is the most challenging and widely used dataset in this domain. Further, we have briefly highlighted the importance of lightweight models that can help various PReID applications. Finally, it is important to consider that beside enhancing the performance of the models, the size of the models (by reducing layers or number of parameters in the model) needs to be reduced without degrading the overall Rank-1 recognition rate.

## References

[1] A. Bedagkar-Gala, S. K. Shah, A survey of approaches and trends in person re-identification, Image and Vision Computing 32 (4) (2014) 270–286.

[2] M. A. Saghafi, A. Hussain, H. B. Zaman, M. H. M. Saad, Review of person re-identification techniques, IET Computer Vision 8 (6) (2014) 455–474.

[3] D. Gray, H. Tao, Viewpoint invariant pedestrian recognition with an ensemble of localized features, in: European conference on computer vision, Springer, 2008, pp. 262–275.

[4] M. Farenzena, L. Bazzani, A. Perina, V. Murino, M. Cristani, Person re-identification by symmetry-driven accumulation of local features, in: Computer Vision and Pattern Recognition (CVPR), 2010 IEEE Conference on, IEEE, 2010, pp. 2360–2367.

[5] M. Hirzer, P. M. Roth, H. Bischof, Person re-identification by efficient impostor-based metric learning, in: Advanced Video and Signal-Based Surveillance (AVSS), 2012 IEEE Ninth International Conference on, IEEE, 2012, pp. 203–208.

[6] B. Ma, Y. Su, F. Jurie, Covariance descriptor based on bio-inspired features for person re-identification and face verification, Image and Vision Computing 32 (6) (2014) 379–390.

[7] A. Bedagkar-Gala, S. K. Shah, A survey of approaches and trends in person re-identification, Image and Vision Computing 32 (4) (2014) 270 – 286. doi:https://doi.org/10.1016/j.imavis.2014.02.001.
URL http://www.sciencedirect.com/science/article/pii/S0262885614000262

[8] Y. LeCun, K. Kavukcuoglu, C. Farabet, Convolutional networks and applications in vision, in: Circuits and Systems (ISCAS), Proceedings of 2010 IEEE International Symposium on, IEEE, 2010, pp. 253–256.

[9] A. Krizhevsky, I. Sutskever, G. E. Hinton, Imagenet classification with deep convolutional neural networks, in: Advances in neural information processing systems, 2012, pp. 1097–1105.

[10] O. Russakovsky, J. Deng, H. Su, J. Krause, S. Satheesh, S. Ma, Z. Huang, A. Karpathy, A. Khosla, M. Bernstein, et al., Imagenet large scale visual recognition challenge, International Journal of Computer Vision 115 (3) (2015) 211–252.

[11] H. O. S. D. Branch, Imagery library for intelligent detection systems (i-lids), in: Crime and Security, 2006. The Institution of Engineering and Technology Conference on, IET, 2006, pp. 445–448.

[12] A. Ess, B. Leibe, L. Van Gool, Depth and appearance for mobile scene analysis, in: Computer Vision, 2007. ICCV 2007. IEEE 11th International Conference on, IEEE, 2007, pp. 1–8.

[13] D. S. Cheng, M. Cristani, M. Stoppa, L. Bazzani, V. Murino, Custom pictorial structures for re-identification., in: BMVC, 2011, p. 6.

[14] W. Li, R. Zhao, X. Wang, Human reidentification with transferred metric learning, in: Asian Conference on Computer Vision, Springer, 2012, pp. 31–44.

[15] W. Li, X. Wang, Locally aligned feature transforms across views, in: Proceedings of the IEEE Conference on Computer Vision and Pattern Recognition, 2013, pp. 3594–3601.

[16] W. Li, R. Zhao, T. Xiao, X. Wang, Deepreid: Deep filter pairing neural network for person re-identification, in: Proceedings of the IEEE Conference on Computer Vision and Pattern Recognition, 2014, pp. 152–159.

[17] M. Hirzer, C. Beleznai, P. M. Roth, H. Bischof, Person re-identification by descriptive and discriminative classification, in: Image Analysis, Springer, 2011, pp. 91–102.

[18] N. Martinel, C. Micheloni, Re-identify people in wide area camera network, in: Computer Vision and Pattern Recognition Workshops (CVPRW), 2012 IEEE Computer Society Conference on, IEEE, 2012, pp. 31–36.

[19] A. Das, A. Chakraborty, A. K. Roy-Chowdhury, Consistent re-identification in a camera network, in: European Conference on Computer Vision, Springer, 2014, pp. 330–345.

[20] L. Zheng, L. Shen, L. Tian, S. Wang, J. Wang, Q. Tian, Scalable person re-identification: A benchmark, in: Proceedings of the IEEE International Conference on Computer Vision, 2015, pp. 1116–1124.

[21] P. F. Felzenszwalb, R. B. Girshick, D. McAllester, D. Ramanan, Object detection with discriminatively trained part-based models, IEEE transactions on pattern analysis and machine intelligence 32 (9) (2010) 1627–1645.

[22] Springer, MARS: A Video Benchmark for Large-Scale Person Re-identification.

[23] E. Ristani, F. Solera, R. Zou, R. Cucchiara, C. Tomasi, Performance measures and a data set for multi-target, multi-camera tracking, in: European Conference on Computer Vision workshop on Benchmarking Multi-Target Tracking, 2016.

[24] L. Wei, S. Zhang, W. Gao, Q. Tian, Person trasfer gan to bridge domain gap for person re-identification, in: Computer Vision and Pattern Recognition, IEEE International Conference on, 2018.

[25] L. Zheng, Y. Yang, A. G. Hauptmann, Person re-identification: Past, present and future, arXiv preprint arXiv:1610.02984.

[26] S. Wu, Y.-C. Chen, X. Li, A.-C. Wu, J.-J. You, W.-S. Zheng, An enhanced deep feature representation for person re-identification, in: Applications of Computer Vision (WACV), 2016 IEEE Winter Conference on, IEEE, 2016, pp. 1–8.

[27] B. Ma, Y. Su, F. Jurie, Local descriptors encoded by fisher vectors for person re-identification, in: European Conference on Computer Vision, Springer, 2012, pp. 413–422.

[28] L. Wu, C. Shen, A. van den Hengel, Deep linear discriminant analysis on fisher networks: A hybrid architecture for person re-identification, Pattern Recognition.

[29] T. Xiao, H. Li, W. Ouyang, X. Wang, Learning deep feature representations with domain guided dropout for person re-identification, in: Proceedings of the IEEE Conference on Computer Vision and Pattern Recognition, 2016, pp. 1249–1258.

[30] C. Su, J. Li, S. Zhang, J. Xing, W. Gao, Q. Tian, Pose-driven deep convolutional model for person re-identification, in: 2017 IEEE International Conference on Computer Vision (ICCV), IEEE, 2017, pp. 3980–3989.

[31] D. Li, X. Chen, Z. Zhang, K. Huang, Learning deep context-aware features over body and latent parts for person re-identification, in: Proceedings of the IEEE Conference on Computer Vision and Pattern Recognition, 2017, pp. 384–393.

[32] G. Zhang, J. Kato, Y. Wang, K. Mase, People re-identification using deep convolutional neural network, in: Computer Vision Theory and Applications (VISAPP), 2014 International Conference on, Vol. 3, IEEE, 2014, pp. 216–223.

[33] D. Yi, Z. Lei, S. Z. Li, Deep metric learning for practical person re-identification, arXiv preprint arXiv:1407.4979.

[34] E. Ahmed, M. Jones, T. K. Marks, An improved deep learning architecture for person re-identification, in: Proceedings of the IEEE Conference on Computer Vision and Pattern Recognition, 2015, pp. 3908–3916.

[35] J. Wang, Z. Wang, C. Gao, N. Sang, R. Huang, Deeplist: Learning deep features with adaptive listwise constraint for person re-identification, IEEE Transactions on Circuits and Systems for Video Technology.

[36] F. Wang, W. Zuo, L. Lin, D. Zhang, L. Zhang, Joint learning of single-image and cross-image representations for person re-identification, in: Proceedings of the IEEE Conference on Computer Vision and Pattern Recognition, 2016, pp. 1288–1296.

[37] H. Liu, J. Feng, M. Qi, J. Jiang, S. Yan, End-to-end comparative attention networks for person re-identification, arXiv preprint arXiv:1606.04404.

[38] S.-Z. Chen, C.-C. Guo, J.-H. Lai, Deep ranking for person re-identification via joint representation learning, IEEE Transactions on Image Processing 25 (5) (2016) 2353–2367.

[39] A. Franco, L. Oliveira, A coarse-to-fine deep learning for person re-identification, in: Applications of Computer Vision (WACV), 2016 IEEE Winter Conference on, IEEE, 2016, pp. 1–7.

[40] A. Franco, L. Oliveira, Convolutional covariance features: Conception, integration and performance in person re-identification, Pattern Recognition 61 (2017) 593–609.

[41] S. Wang, C. Zhang, L. Duan, L. Wang, S. Wu, L. Chen, Person re-identification based on deep spatio-temporal features and transfer learning, in: Neural Networks (IJCNN), 2016 International Joint Conference on, IEEE, 2016, pp. 1660–1665.

[42] H. Shi, Y. Yang, X. Zhu, S. Liao, Z. Lei, W. Zheng, S. Z. Li, Embedding deep metric for person re-identification: A study against large variations, in: European Conference on Computer Vision, Springer, 2016, pp. 732–748.

[43] N. McLaughlin, J. M. del Rincon, P. Miller, Person re-identification using deep convnets with multi-task learning, IEEE Transactions on Circuits and Systems for Video Technology.

[44] S. Iodice, A. Petrosino, I. Ullah, Strict pyramidal deep architectures for person re-identification, in: Advances in Neural Networks, Springer, 2016, pp. 179–186.

[45] J. Zhu, H. Zeng, S. Liao, Z. Lei, C. Cai, L. Zheng, Deep hybrid similarity learning for person re-identification, arXiv preprint arXiv:1702.04858.

[46] D. Tao, Y. Guo, B. Yu, J. Pang, Z. Yu, Deep multi-view feature learning for person re-identification, IEEE Transactions on Circuits and Systems for Video Technology.

[47] S. Liao, Y. Hu, X. Zhu, S. Z. Li, Person re-identification by local maximal occurrence representation and metric learning, in: Proceedings of the IEEE Conference on Computer Vision and Pattern Recognition, 2015, pp. 2197–2206.

[48] C. Mao, Y. Li, Z. Zhang, Y. Zhang, X. Li, Pyramid person matching network for person re-identification, in: Asian Conference on Machine Learning, 2017, pp. 487–497.

[49] R. R. Varior, B. Shuai, J. Lu, D. Xu, G. Wang, A siamese long short-term memory architecture for human re-identification, in: European Conference on Computer Vision, Springer, 2016, pp. 135–153.

[50] X. Qian, Y. Fu, Y.-G. Jiang, T. Xiang, X. Xue, Multi-scale deep learning architectures for person re-identification, arXiv preprint arXiv:1709.05165.

[51] S. Ding, L. Lin, G. Wang, H. Chao, Deep feature learning with relative distance comparison for person re-identification, Pattern Recognition 48 (10) (2015) 2993–3003.

[52] R. Zhang, L. Lin, R. Zhang, W. Zuo, L. Zhang, Bit-scalable deep hashing with regularized similarity learning for image retrieval and person re-identification, IEEE Transactions on Image Processing 24 (12) (2015) 4766–4779.

[53] D. Cheng, Y. Gong, S. Zhou, J. Wang, N. Zheng, Person re-identification by multi-channel parts-based cnn with improved triplet loss function, in: Proceedings of the IEEE Conference on Computer Vision and Pattern Recognition, 2016, pp. 1335–1344.

[54] C. Su, S. Zhang, J. Xing, W. Gao, Q. Tian, Deep attributes driven multi-camera person re-identification, in: European Conference on Computer Vision, Springer, 2016, pp. 475–491.

[55] J. Liu, Z.-J. Zha, Q. Tian, D. Liu, T. Yao, Q. Ling, T. Mei, Multi-scale triplet cnn for person re-identification, in: Proceedings of the 2016 ACM on Multimedia Conference, ACM, 2016, pp. 192–196.

[56] D. Cheng, Y. Gong, Z. Li, W. Shi, A. G. Hauptmann, N. Zheng, Deep feature learning via structured graph laplacian embedding for person re-identification, arXiv preprint arXiv:1707.07791.

[57] X. Bai, M. Yang, T. Huang, Z. Dou, R. Yu, Y. Xu, Deep-person: Learning discriminative deep features for person re-identification, arXiv preprint arXiv:1711.10658.

[58] J. Lin, L. Ren, J. Lu, J. Feng, J. Zhou, Consistent-aware deep learning for person re-identification in a camera network, in: The IEEE Conference on Computer Vision and Pattern Recognition (CVPR), Vol. 6, 2017.

[59] R. Hadsell, S. Chopra, Y. LeCun, Dimensionality reduction by learning an invariant mapping, in: Computer vision and pattern recognition, 2006 IEEE computer society conference on, Vol. 2, IEEE, 2006, pp. 1735–1742.

[60] L. Wu, C. Shen, A. v. d. Hengel, Personnet: Person re-identification with deep convolutional neural networks, arXiv preprint arXiv:1601.07255.

[61] N. McLaughlin, J. M. del Rincon, P. Miller, Recurrent convolutional network for video-based person re-identification, in: Computer Vision and Pattern Recognition (CVPR), 2016 IEEE Conference on, IEEE, 2016, pp. 1325–1334.

[62] Y. Bengio, Rmsprop and equilibrated adaptive learning rates for non-convex optimization, corr abs 1502.

[63] B. Prosser, W.-S. Zheng, S. Gong, T. Xiang, Q. Mary, Person re-identification by support vector ranking., in: BMVC, 2010, p. 6.

[64] H. Shi, X. Zhu, S. Liao, Z. Lei, Y. Yang, S. Z. Li, Constrained deep metric learning for person re-identification, arXiv preprint arXiv:1511.07545.

[65] S. Li, X. Liu, W. Liu, H. Ma, H. Zhang, A discriminative null space based deep learning approach for person re-identification, in: Cloud Computing and Intelligence Systems (CCIS), 2016 4th International Conference on, IEEE, 2016, pp. 480–484.

[66] D. Chung, K. Tahboub, E. J. Delp, A two stream siamese convolutional neural network for person re-identification, in: Proceedings of the IEEE Conference on Computer Vision and Pattern Recognition, 2017, pp. 1983–1991.

[67] B. Lavi, G. Fumera, F. Roli, Multi-stage ranking approach for fast person re-identification, IET Computer Vision.

[68] B. Lavi, G. Fumera, F. Roli, A multi-stage approach for fast person re-identification, in: Joint IAPR International Workshops on Statistical Techniques in Pattern Recognition (SPR) and Structural and Syntactic Pattern Recognition (SSPR), Springer, 2016, pp. 63–73.